\documentclass[10pt,twocolumn,letterpaper]{article}
\usepackage{silence}
\usepackage[accsupp]{axessibility}
\usepackage{cvpr} 
\usepackage{xcolor}
\definecolor{darkred}{rgb}{0.8, 0.0, 0.0}
\definecolor{orange}{rgb}{1.0, 0.347, 0.0}

\newcommand{\taskname}{\textbf{FloVerse}\xspace}
\newcommand{\policyname}{\textbf{ThreeDiff}\xspace}
\newcommand{\datasetname}{\textbf{\taskname-$1.6$K}\xspace}
\newcommand{\modelA}{\textbf{planner}\xspace}
\newcommand{\modelB}{\textbf{refiner}\xspace}
\newcommand\supp{\textit{supplementary material}\xspace}

\usepackage[utf8]{inputenc}
\usepackage[T1]{fontenc}
\usepackage{latexsym}
\usepackage{microtype}
\usepackage{booktabs}
\usepackage{graphicx}
\usepackage{algorithmic}
\usepackage[linesnumbered,ruled,vlined]{algorithm2e}
\usepackage{acronym}
\usepackage{balance}
\usepackage[skip=3pt,font=small]{caption}
\usepackage{subcaption}
\usepackage{xspace}
\usepackage{setspace}
\usepackage{tabularx,colortbl,multirow,array,makecell}
\usepackage{overpic,wrapfig}
\usepackage{nicefrac}
\usepackage[misc]{ifsym}
\usepackage{siunitx}
\usepackage[dvipsnames,svgnames,x11names]{xcolor}
\usepackage{soul} 
\usepackage{pifont}  
\usepackage[pagebackref,breaklinks,colorlinks,citecolor=cvprblue]{hyperref}
\usepackage[capitalize]{cleveref}
\usepackage{lipsum}
\usepackage{bm}
\usepackage{xcolor}

\acrodef{eai}[EAI]{Embodied AI}
\acrodef{s2r}[Sim2Real]{sim-to-real}
\acrodef{vln}[VLN]{vision-language navigation}
\acrodef{mcmc}[MCMC]{Markov-Chain Monte-Carlo}
\acrodef{cd}[CD]{Chamfer Distance}
\acrodef{ecd}[ECD]{Enhanced Chamfer Distance}
\acrodef{bbox}[Bbox]{Bounding box}
\acrodef{iou}[IoU]{Intersection over Union}
\acrodef{agv}[AGV]{Automated Guided Vehicle}
\acrodef{dwa}[DWA]{Dynamic Window Approach}

\definecolor{WeiColor}{rgb}{1,0.33,0.64}
\definecolor{myDarkBlue}{RGB}{55,81,139}
\definecolor{myLightBlue}{RGB}{158,186,217}
\definecolor{myLightGreen}{RGB}{126,171,85}
\definecolor{myRed}{RGB}{176,36,24}
\definecolor{JxColor}{RGB}{50,102,204}
\definecolor{LightGray}{gray}{0.9}
\definecolor{cvprblue}{rgb}{0.21,0.49,0.74}
\definecolor{softlime}{RGB}{180, 229, 162}

\definecolor{pointcolor}{RGB}{233,113,50}   % Orange
\definecolor{objectcolor}{RGB}{15,158,213}  % Blue
\definecolor{imagecolor}{RGB}{78,167,46}    % Green

\makeatletter
\DeclareRobustCommand\onedot{\futurelet\@let@token\@onedot}
\def\@onedot{\ifx\@let@token.\else.\null\fi\xspace}

\makeatother

\makeatletter
\renewcommand{\paragraph}{%
  \@startsection{paragraph}{4}%
  {\z@}{0ex \@plus 0ex \@minus 0ex}{-1em}%
  {\hskip\parindent\normalfont\normalsize\bfseries}%
}
\makeatother

\newcommand\blfootnote[1]{%
  \begingroup
  \renewcommand\thefootnote{}\footnote{#1}%
  \addtocounter{footnote}{-1}%
  \endgroup
}

\def\paperID{} 
\def\confName{CVPR}
\def\confYear{2026}

\title{FloVerse: Floor Plan-Guided Multi-Modal Navigation}
\author{%
    Weiqi Huang$^{1}$, Shuangyi Dong$^{1}$, Jiaxin Li$^{1}$, Yifei Guo$^{1}$, Zan Wang$^{1}$, Wei Liang$^{1\,\textrm{\Letter}}$
    \vspace{6pt}\\
    \small $^1$ School of Computer Science \& Technology, Beijing Institute of Technology
    \vspace{6pt}\\
    \href{https://wikiahuang.github.io/floverse/}{https://wikiahuang.github.io/floverse/}
}
\begin{document}

\twocolumn[{%
\renewcommand\twocolumn[1][]{#1}%
\maketitle
\vspace{-15pt}
\begin{center}
    \centering
    \captionsetup{type=figure}
    \includegraphics[width=\linewidth]{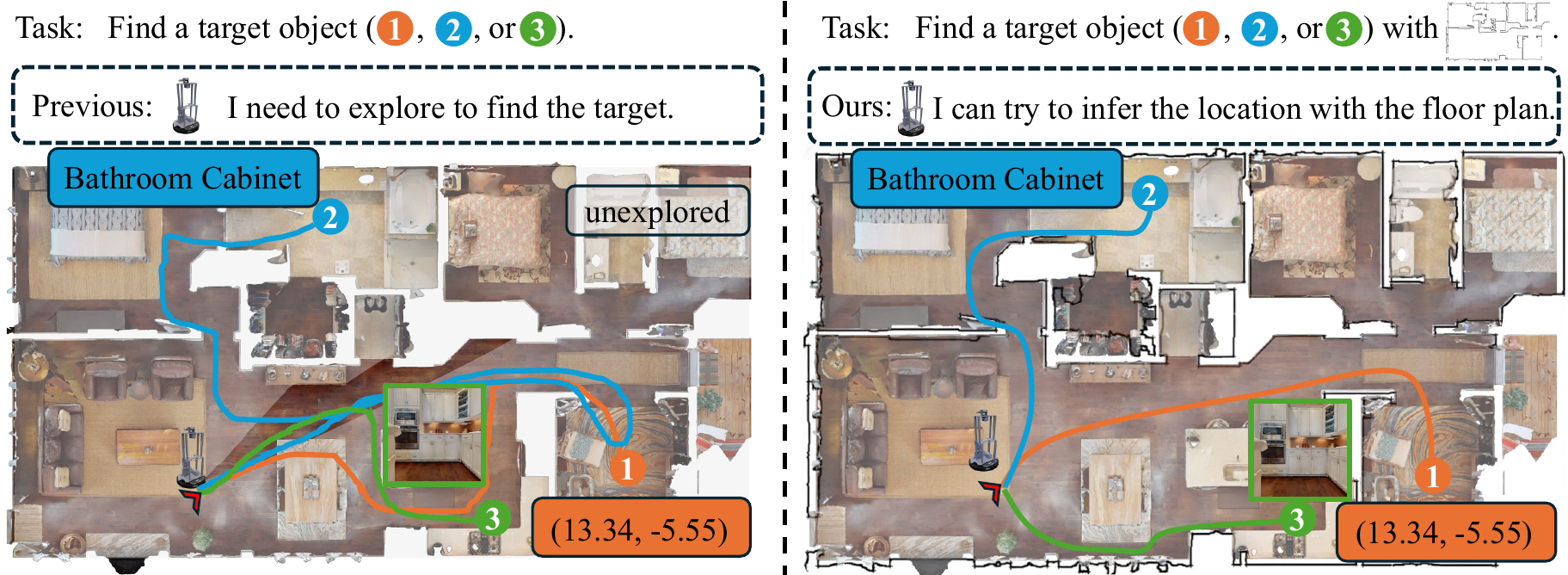}
    \captionof{figure}{
        \textbf{Overview of \taskname.} The agent is given a goal of any modality (\textcolor{pointcolor}{PointNav}, \textcolor{objectcolor}{ObjectNav}, or \textcolor{imagecolor}{ImageNav}). Prior methods rely only on local observations and must explore unseen areas. In contrast, \policyname leverages floor-plan spatial priors for more efficient navigation.
    }
    \label{fig:teaser}
\end{center}
}]

\blfootnote{$^\textrm{\Letter}$ Corresponding author: Wei Liang (\texttt{liangwei@bit.edu.cn}).}

\begin{abstract}%
Floor plans encapsulate compact spatial priors, enabling agents to navigate unseen scenes more efficiently.
While prior work has explored floor plan–guided navigation, it has focused mainly on PointNav and a limited set of environments.
To bridge this gap, we introduce \taskname, a new task for floor plan–guided embodied navigation that unifies PointNav, ObjectNav, and ImageNav. 
To support this \taskname, we assemble \datasetname, a large-scale dataset of $1.6$K scenes from HM3D and Gibson $4$+, paired with corresponding floor plans, comprising 240K expert trajectories and 12M RGBD frames. 
We further propose \policyname, a two-stage imitation learning policy comprising a \modelA, a diffusion-based multimodal goal-reasoning module trained via masked-modality modeling, and a \modelB, a depth-based trajectory-refinement module for safe execution.
Extensive experiments demonstrate that 
(1) floor-plan priors improve navigation performance across all goal modalities, and 
(2) \policyname implicitly captures spatial information from floor plans. These results underscore the effectiveness of spatial priors and validate our proposed unified approach for floor plan–guided embodied navigation.
% Our project website is \href{https://wikiahuang.github.io/floverse/}{https://wikiahuang.github.io/floverse/}.
% \vspace{-6pt}%
\end{abstract}%

\section{Introduction}
\label{sec:intro}
Embodied navigation is advancing at an unprecedented pace, achieving impressive performance across challenging tasks such as point-goal (PointNav) \cite{anderson2018evaluation, roth2024viplanner, wijmans2020ddppo, zhao2021surprising}, object-goal (ObjectNav) \cite{chaplot2020object, majumdar2022zson, huang2024gamap, yin2024sg, yokoyama2024vlfm}, and image-goal (ImageNav) \cite{krantz2023navigating, sridhar2023nomad, sun2023fgprompt} navigation.
Despite these achievements, most methods still depend on active mapping or extensive exploration, which often leads to short-sighted behaviors that reduce overall efficiency (\cref{fig:teaser}).

Recent studies \cite{li2024flona, ewe2024spatial} explore using floor plans as spatial priors to guide navigation \cite{ewe2024spatial, li2024flona}. Floor plans are easily accessible and encode informative geometric structure,
% rich geometric layout, 
making them inherently well-suited for navigation in unseen environments.
However, existing studies focus mainly on PointNav and limited scenes, leaving the broader potential of floor plan–guided navigation largely unexplored.

In fact, floor plans contain not only geometric structures but also implicit semantic information about the environment layout, such as room functionalities and typical object distributions.
These geometric and semantic regularities may provide rich spatial priors that enable agents to infer goal locations and plan more effectively, potentially enhancing embodied navigation across diverse goal modalities.

To investigate how floor plan priors can facilitate embodied navigation, we introduce \taskname, a new task in which an agent must navigate to a goal—specified as a point, object, or image—using its egocentric RGBD observations and a floor plan in unseen environments (\cref{fig:teaser}).
To support this task, we develop a fully automated pipeline that derives floor plans from 3D scene meshes, producing a diverse collection of $1,627$ floor plans across HM3D \cite{ramakrishnan2021hm3d} and Gibson $4$+ \cite{xia2018gibson, savva2019habitat} datasets. Note that each floor plan corresponds to a single level of a scene.
We then leverage SpatialLM \cite{SpatialLM} to enrich object annotations in HM3D scenes, enhancing both scene coverage and category diversity. After manual verification, the resulting dataset contains $325$ object categories across $299$ scenes. Finally, we collect expert navigation trajectories across all three modalities, yielding $240$K waypoint sequences and over $12$M RGBD–pose pairs.
Detailed statistics and the collection pipeline are provided in \cref{sec:dataset}.

Building on this task, we propose \policyname, a unified two-stage end-to-end policy that exploits both global floor plan spatial priors and local geometric cues to efficiently navigate across diverse goal modalities.
In the first stage, a diffusion-based \modelA predicts a goal-conditioned coarse trajectory, capturing high-level spatial intent and long-horizon dependencies.
In the second stage, the \modelB, a local refinement module, utilizes depth-derived occupancy cues to adapt the coarse trajectory to the agent’s immediate surroundings, generating collision-free motions.

We extensively evaluate \policyname on \taskname to assess the impact of incorporating floor plan priors.
Experimental results show that floor plan information consistently improves navigation efficiency and success rates across all goal modalities.
Beyond quantitative gains, \policyname exhibits an emergent ability to infer goals' location, even without explicit supervision.
Compared to specialized baselines trained for individual goal modalities, \policyname delivers comparable or superior performance, highlighting the effectiveness and generality of the proposed unified framework.

Our main contributions are:
\begin{itemize}
    \item \textbf{A multi-modal floor plan–guided navigation task.} \taskname introduces navigation to point, object, or image goals using floor plans and egocentric observations.
    \item \textbf{A supporting dataset, \datasetname.} It offers a rich set of reconstructed floor plans and expert navigation trajectories spanning a diverse range of scenes.
    \item \textbf{A two-stage navigation model.} \policyname integrates global planning with local obstacle awareness, achieving strong performance.
    \item \textbf{Empirical validation of floor plan priors.} Quantitative and qualitative analysis show that incorporating floor plans consistently improves navigation, highlighting their value as structured spatial priors.
\end{itemize}

\section{Related Work}
\label{sec:related work}
\subsection{Visual Navigation}
Current visual navigation methods generally fall into two categories: mapping-based and mapless approaches. 
The former typically integrates semantic information into various map representations. Normally, researchers extract semantic categories \cite{cha2020obj, ram2022pon, guo2024object, yu2024trajectory}, pixel-level features \cite{huang2023visual, yokoyama2024vlfm}, and implicit image features \cite{pan2025planning, chen2022think, wang2023dreamwalker} to construct comprehensive maps or topological graphs that support subsequent planning. 
In contrast, mapless visual navigation methods focus on directly translating observations into actions. 
Previous work has demonstrated notable success in PointNav through large-scale trial-and-error reinforcement learning \cite{wijmans2020ddppo}. % and learning-based visual odometry \jx{visual odometry???} \cite{zhao2021surprising, partsey2022ismapping}. 
Nomad \cite{sridhar2023nomad} utilizes a diffusion policy to generate smooth trajectories for exploration and image-goal navigation. 
More recently, an increasing number of studies have leveraged the reasoning capabilities of large language models (LLMs) to guide agents in navigation \cite{Zheng_2024_CVPR, chen2025affordances, zhou2024navgpt, cao2024cognav, xu2025mm}. 
However, mapping-based methods incur significant overhead from explicit map construction, whereas mapless methods lack global guidance when the goal is not observable. Both limitations hinder navigation efficiency.

\subsection{Floor Plan Guided Navigation}
\label{subsec:Floorplan Navigation}
Given the inherent limitations of the aforementioned methods, recent research has begun to explore floor plan-guided visual navigation. 
In indoor environments, floor plans provide readily available spatial priors that facilitate efficient visual navigation. 
However, the absence of detailed information about movable objects, such as furniture, makes floor plans temporally stable but also presents challenges for directly planning collision-free trajectories.
To address this limitation, prior studies have incorporated Voronoi diagrams \cite{setalaphruk2003robot}, pre-collected visual features \cite{li2021cognitive}, traversability cues from robot observations \cite{ewe2024spatial}, and 2D point cloud data \cite{goswami2024floor} to enhance floor plan representations and improve localization and navigation. FloNa \cite{li2024flona} makes the first attempt to navigate in unseen environments using only a monocular camera and a floor plan within an end-to-end framework. 

Although floor plans have been shown to improve navigation efficiency, prior evaluations have been limited to small-scale environments and the PointNav task. Their impact on other goal modalities, such as ImageNav and ObjectNav, is still largely unexplored. In this work, we scale up the evaluation and systematically examine whether floor plans can enhance navigation performance across PointNav, ObjectNav, and ImageNav.

\subsection{Multi-Modal Goal-Oriented Navigation}
\label{subsec:Multi-Modal Navigation}
Early navigation models typically address a single goal modality. More recently, research has shifted toward developing unified frameworks capable of handling multiple goal types within a single model.
For instance, VIENNA \cite{wang2022towards} unifies four navigation tasks through an attention-based parse-and-query framework, demonstrating that a multi-task agent can match or surpass task-specific models. Goat-Bench \cite{khanna2024goat} introduces a multi-modal lifelong navigation setting, where an agent processes goals expressed as images, object categories, and textual descriptions using a modular skill-chain architecture. Uni-Goal \cite{yin2025unigoal} represents all goal modalities as a unified graph structure and leverages an online scene graph to infer goal locations dynamically. NavDP \cite{cai2025navdp} adopts a diffusion-based formulation, encoding goal types as modality tokens fused with RGBD inputs for joint policy learning. OmniNav \cite{Xue2025OmniNavAU} features a fast-slow collaborative architecture and large-scale multi-task training to achieve strong generalization across diverse navigation tasks.
Building on these advances, we introduce the first model that unifies three goal modalities—PointNav, ObjectNav, and ImageNav—within a floor plan–guided navigation framework.

\section{Dataset}
\label{sec:dataset}
\subsection{Floor Plan Construction}
\begin{figure}[t!]
    \centering
    \includegraphics[width=\linewidth]{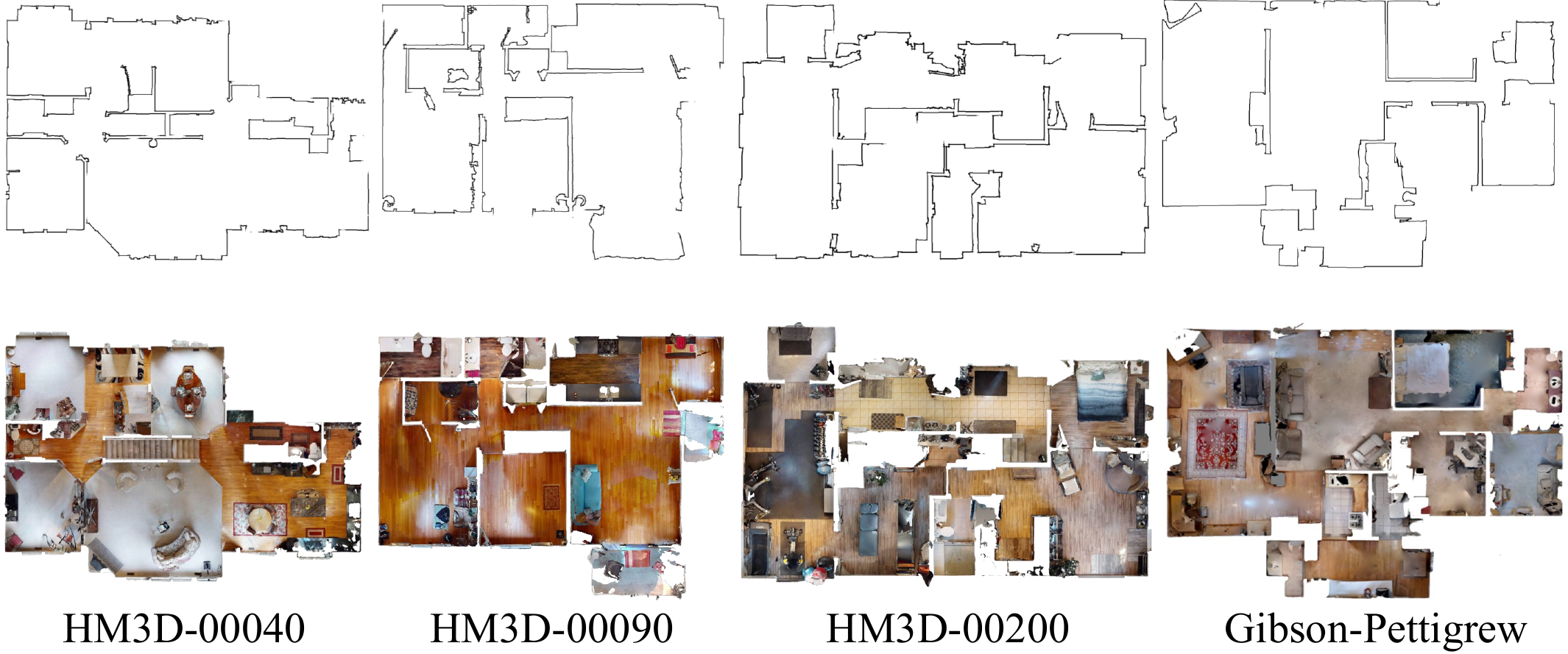}
    \caption{\textbf{Examples of constructed floor plans.} Top row: generated floor plans. Bottom row: corresponding 3D scenes from the HM3D and Gibson $4$+ datasets}
    \label{fig:floorplan_examples}
\end{figure}

To support \taskname, we extract stable structural elements, such as walls, from object-containing meshes, as these elements correspond to features represented in floor plans. Vertices are first filtered based on their height and normal orientation, with a threshold of $1.25\mathrm{m}$, effectively removing most objects. To further mitigate reconstruction artifacts and spurious noise, we apply morphological operations, including dilation and erosion, for denoising. Using this pipeline, we construct a total of $1,627$ floor plans, comprising $1,488$ from the HM3D scenes and $139$ from Gibson $4$+ scenes. Examples of the resulting floor plans are shown in \cref{fig:floorplan_examples}.

\subsection{Object Goal Annotation}
\label{subsec: obj annotation}
\begin{figure}[t!]
    \centering
    \includegraphics[width=\linewidth]{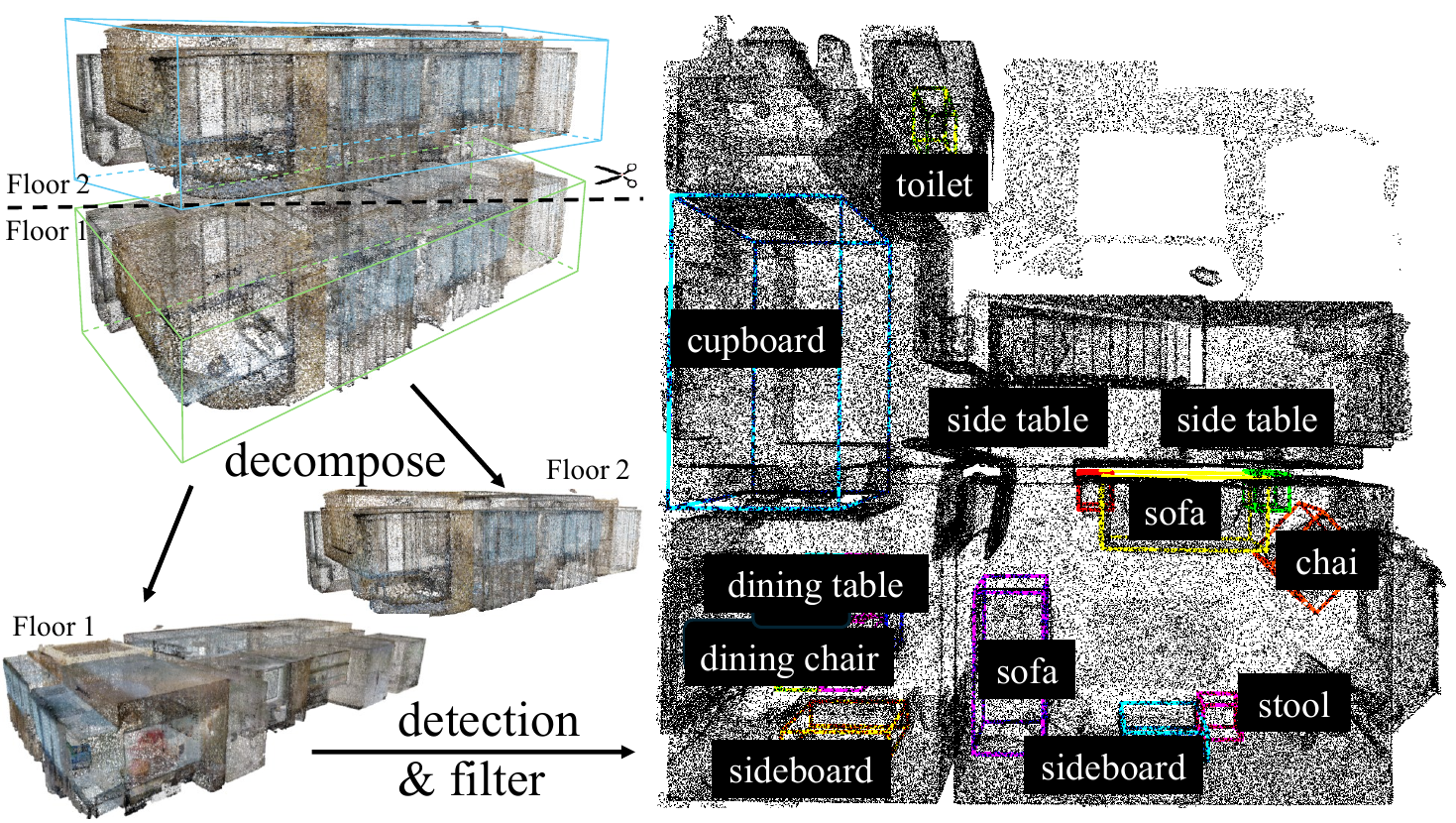}
    \caption{\textbf{Our object annotation pipeline.} Each scene point cloud is first decomposed into separate layers and then processed by SpatialLM for object detection. The detected objects are manually filtered to remove incorrect annotations, yielding the final results.}
    \label{fig:object_annotation}
\end{figure}

Most object-goal navigation methods involve a limited range of object categories, typically $6$ \cite{cha2020obj} or $21$ \cite{batra2020objectnav}. HM3D-OVON \cite{yokoyama2024hm3d} substantially expands the coverage to $379$ categories but includes only $181$ scenes. We further clean this dataset by removing objects whose associated viewpoints fall outside navigable regions, resulting in $304$ object categories across $169$ scenes.
To enrich category and scene diversity, we employ SpatialLM \cite{SpatialLM} for additional object recognition, and manually filter annotations to ensure quality.
As illustrated in \cref{fig:object_annotation}, SpatialLM is applied to each scene point cloud to detect objects and extract their bounding boxes and poses. For multi-layer scenes, we perform layer-wise decomposition to maintain accuracy. In total, \datasetname contains $325$ object categories spanning $299$ scenes.

\begin{table}[h]
    \centering
    \caption{\textbf{Data splits and the number of navigation episodes.} IO episodes refer to the episodes supporting ImageNav ObjectNav.}
    \label{tab:datasets}
    \resizebox{\linewidth}{!}{%
        \begin{tabular}{ccccc}%
            \toprule
            \multicolumn{1}{c}{\multirow{2}[1]{*}{Dataset}} & \multicolumn{2}{c}{train} & \multicolumn{2}{c}{eval} \\ 
            \cmidrule(rr){2-3}\cmidrule(lr){4-5}
                             &  HM3D & Gibson $4$+ & HM3D  & Gibson $4$+ \\ 
            \midrule
            scenes           & $1321$    & $121$   & $167$    & $18$ \\ 
            total episodes   & $198,150$ & $18,150$ & $25,050$ & $2,700$\\ 
            IO episodes      & $65,700$  &    -    & $8,550$  & - \\ 
            % object category &  &  &  &  \\ 
            \bottomrule
        \end{tabular}%
    }%
\end{table}

\subsection{Trajectory Collection}
We construct a large-scale set of expert trajectories covering all goal modalities for both imitation learning and evaluation. For PointNav, goals are randomly sampled positions on the navigable map (details of navigability are provided in the \supp), whereas for ObjectNav and ImageNav, goals correspond to annotated object locations. The start position is randomly chosen from navigable areas at least $5\mathrm{m}$ away from the goal. The shortest path between each start–goal pair is computed using the A* algorithm and discretized into waypoints at $10\mathrm{cm}$ intervals. At each waypoint, the agent’s pose and RGBD observations are recorded.
Each expert trajectory includes RGBD–pose pairs, a floor plan, and a goal. The goal is defined as a point in PointNav, an image in ImageNav, and an object category in ObjectNav. As summarized in \cref{tab:datasets}, the dataset contains about $240$K trajectories with $12$M RGBD–pose pairs, including roughly $74$K ImageNav/ObjectNav episodes (IO episodes).

\section{Problem Definition}
\taskname aims to enable floor plan–guided embodied navigation in unseen indoor environments.
Formally, given an environment $\mathcal{E}$ that provides both perceptual observations and a global 2D floor plan $F$, the agent interacts with $\mathcal{E}$ through a sequence of observations, actions, and state transitions.
Specifically, at each time step $t$, the agent receives a state $\mathbf{s}_t = (I_t, D_t, p_t)$, where $I_t \in \mathbb{R}^{H \times W \times 3}$ and $D_t \in \mathbb{R}^{H \times W}$ denote the egocentric RGB and depth images, and $\mathbf{p}_t = (x_t, y_t, \theta_t)$ represents the agent’s 2D pose in the environment.
The floor plan $F \in \mathbb{R}^{H_f \times W_f}$ serves as a spatial prior describing the structural layout of $\mathcal{E}$. The goal $g$ depends on the navigation modality:
\begin{equation}
    g =
    \begin{cases}
    g_{\text{point}} \in \mathbb{R}^2, & \text{PointNav} \\
    g_{\text{object}} \in \mathcal{L}, & \text{ObjectNav} \\
    g_{\text{image}} \in \mathbb{R}^{H \times W \times 3}, & \text{ImageNav}
    \end{cases}
\end{equation}
where $\mathcal{L}$ is the set of textual object labels.

At each step, the agent executes an action $a_t \in \mathcal{A}$ and transitions according to
\begin{equation}
    p_{t+1}=Tr(\mathcal{E}, p_t, a_t),
\end{equation}
which updates its state and generates a new observation $\mathbf{s}_{t+1}$. The action space $\mathcal{A}$ is either a discrete set of primitives $\mathcal{A}_{\text{discrete}} = \{\textit{move forward}, \textit{turn left}, \textit{turn right}, \textit{stop}\}$,  
or a continuous set of 2D waypoints $\mathcal{A}_{\text{continuous}} \subset \mathbb{R}^2$.
An episode is represented as $\tau = {(\mathbf{s}_t, a_t)}_{t=1}^{T}$, where $T$ is the episode length.
An episode is considered successful if the agent reaches the goal $g$ within a distance threshold $d$ before exceeding the maximum number of steps $T_{\max}$ or collisions $\delta_{\max}$.
The objective is to learn a policy $\pi(a_t \mid \mathbf{s}_t, F, g)$ that efficiently drives the agent to the goal while minimizing trajectory length and collisions.

\section{Method}
\begin{figure*}[t!]
    \centering
    \includegraphics[width=\linewidth]{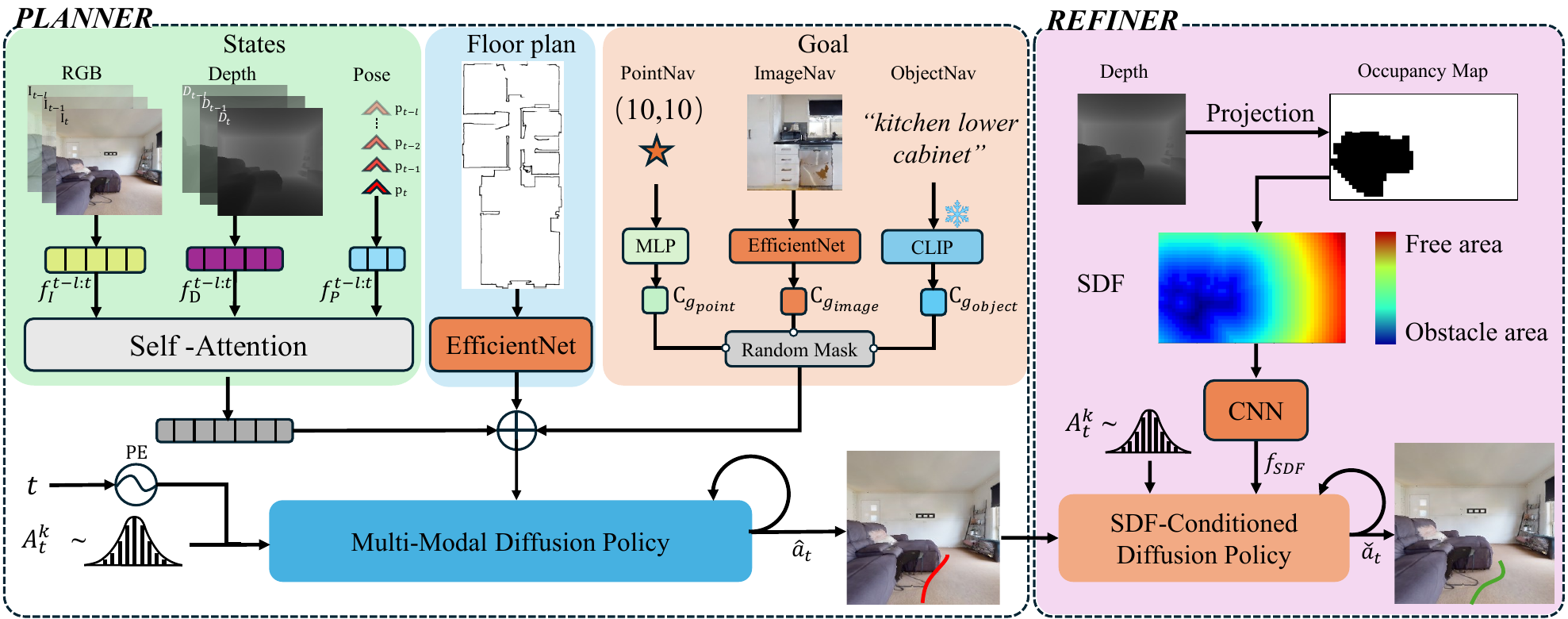}
    \caption{\textbf{Overview of \policyname.} \policyname is a two-stage trajectory generation framework for floor plan–guided navigation. In the first stage, goal-related features, floor plan representations, and observation embeddings are concatenated and used as the conditioning input to a diffusion model, which generates a coarse initial trajectory. In the second stage, depth-derived obstacle-awareness features are integrated with the initial trajectory and fed into a second diffusion model, which refines the trajectory to yield a final, obstacle-aware navigation path.}
    \label{fig:framework}
\end{figure*}

We introduce a two-stage navigation model that generates obstacle-aware trajectories for all three goal modalities using the agent’s egocentric observation and the floor plan. As shown in \cref{fig:framework}, the \modelA first predicts a coarse trajectory using the floor plan and goal information, where the goal modality is randomly masked during training to encourage modality-agnostic reasoning. The \modelB then refines this trajectory using local geometric cues from the depth observation, producing a final path that avoids nearby obstacles.

\subsection{Multi-Modality Goal Conditioned Planner}
To enable a single model to support diverse goal modalities, we employ a random modality masking strategy during training. 
At each iteration, the model receives only one of the three goal modalities, while the other two are masked. The resulting masked goal condition is defined as:
\begin{equation}
    C_{g} = \mathbf{m} \circ (C_{g_{\text{point}}}, C_{g_{\text{object}}}, C_{g_{\text{image}}})
\end{equation}
where $\mathbf{m} = [m_{\text{point}}, m_{\text{object}}, m_{\text{image}}]$ is a binary vector with exactly one entry set to $1$.

The point goal condition is encoded using an MLP:
\begin{equation}
C_{g_{\text{point}}} = \text{MLP}(g_{\text{point}}),
\end{equation}
where $\text{MLP}(\cdot)$ denotes a multilayer perceptron.
For ImageNav, we obtain the image-goal condition by encoding both the current RGB observation $I_t$ and the goal image $g_{\text{image}}$ with an image encoder $\text{E}(\cdot)$, instantiated as EfficientNet, followed by feature concatenation:
\begin{equation}
    C_{g_{\text{image}}} = \text{MLP}(\text{E}(I_t) \oplus \text{E}(g_{\text{image}})). 
\end{equation}
Similarly, the object-goal condition is obtained using a CLIP encoder $\text{CLIP}(\cdot)$ applied to the current image and the target object label:
\begin{equation}
    C_{g_{\text{object}}} = \text{MLP} (\text{CLIP}(I_t) \oplus \text{CLIP}(g_{\text{object}})).
\end{equation}

This strategy provides two advantages.
First, by exposing the model to all three modalities under the same input setting $(s_t, F, \bar{a}_t)$ across training iterations, where $\bar{a}_t$ represents the ground truth action, the modalities effectively inform and reinforce one another, leading to more generalizable trajectory generation.
Second, by always combining the goal with features from the current observation, the model learns a consistent notion of current–goal correspondence across modalities, thereby improving stability and sample efficiency during diffusion training.

Then, we introduce a diffusion policy to learn a goal-oriented floor plan-guided navigation policy that takes $C_{g}$, $C_{\mathcal{F}}$, and $C_{\mathcal{O}}$ as inputs to generate the current action $\hat{a}_t$. Here, $C_{\mathcal{F}}$ denotes floor plan features extracted using EfficientNet, and $C_{\mathcal{O}}$ represents the current observation features, which is computed via a multi-head self-attention module ($SA$): $C_{\mathcal{O}} = SA(f^{t-l:t}_I \oplus f^{t-l:t}_D \oplus f^{t-l:t}_p)$, where $ f^{t-l:t}_I$, $ f^{t-l:t}_D$, and $ f^{t-l:t}_p$ represent the RGB, depth, and pose features from the past $ l$ timesteps, respectively.

To model $ P_{\theta}(\bar{a}_t | C_{g}, C_{\mathcal{F}}, C_{\mathcal{O}}) $, we deploy a conditional U-Net \cite{ronneberger2015u} and DDPM \cite{ho2020denoising} scheduler to perform the forward and backward process. The training objective is:
\begin{equation}
    \mathcal{L}_1 = \text{MSE}(\epsilon_k, \epsilon_{\theta}(\bar{a}_t + \epsilon_k, k)),
\end{equation} 
where MSE represents the mean squared error, $\epsilon$ denotes the noise label, and $k$ represents the denoising steps.

\subsection{Depth-Based Trajectory Refiner}
The training trajectories are optimal, collision-free paths, meaning the model does not receive collision feedback, which limits its ability to actively avoid obstacles. Since obstacle avoidance is fundamentally tied to understanding spatial geometry, we address this limitation by training a second diffusion model for safer trajectory prediction using the geometric information from depth. As shown in \cref{fig:framework}, we first project the depth into a local binary occupancy map from a Bird's Eye View (BEV) perspective, with a grid size of $20 \times 50$ and a resolution of $0.1\mathrm{m}$, where $0$ represents obstacles and $1$ represents free space. Next, we compute a local Signed Distance Field (SDF), which is then encoded using a CNN. The resulting embedding $f_{SDF}$ is combined with the predicted trajectory $\hat{a}_t$ and fed into the second diffusion model, which outputs the refined trajectory 
 $\check{a}_t$. The diffusion process aims to fit the following distribution: $ P_{\phi}(\check{a}_t | \hat{a}_t, f_{SDF}). $

In addition to minimizing the discrepancy between the predicted and true trajectories, $ \check{a}_t$ should also be as distant as possible from obstacles. Thus, the training objective is defined as:
\begin{equation}
    \mathcal{L}_2 = \text{MSE}(\epsilon_k, \epsilon_{\phi}(\overline{a}_t + \epsilon_k, k)) + \alpha \mathcal{L}_{\text{collision}}, 
\end{equation}
and $\mathcal{L}_{\text{collision}} = \frac{1}{N} \sum_{i=1}^{N} \exp\left( - \lambda \cdot \rho_i \right)$, where $N$ is the number of waypoints,  $\rho_i$ is the $l_2$ distance from the $i$-th point to the nearest obstacle, where  $\alpha$ and $\lambda$ are hyperparameters that controls the weight of the collision loss. Note that, to ensure differentiability, $\rho_i$ is computed from the SDF after applying linear interpolation.

\subsection{Implement Details}
In our implementation, we first train the first diffusion model until convergence, and then integrate the second diffusion model, training both jointly until the system reaches convergence. For the object goal condition, we utilize a frozen CLIP-ViT-B/$32$ model to encode the images and text labels. All other image encoders in \policyname are EfficientNet-B0 with non-shared weights, trained from scratch. The multi-head attention mechanism consists of four heads, each with four layers. We optimize the models using the AdamW optimizer, with the learning rate governed by PyTorch's CosineAnnealingLR scheduler. The maximum learning rate is set to $0.0001$, and training is conducted for a maximum of $20$ epochs. The parameter 
$\alpha$ is set to $0.1$, and $\lambda$ is set to $1$. The training of \policyname is performed on $4$ NVIDIA $4090$ GPUs, with a batch size of $32$ per GPU.

\section{Experiments}
In this section, we examine four key aspects: (1) the impact of floor plans on performance in PointNav, ImageNav, and ObjectNav (\cref{subsec: impact of floorplan}); (2) the cross-modality complementarity effect under random modality masking (\cref{subsec: random masking}); (3) the performance of existing navigation models on FloVerse (\cref{subsec: comparison}); and (4) the contribution of the depth-based refiner to the overall performance of ThreeDiff (\cref{subsection: second stage}).
\subsection{Setup and Metrics}
\paragraph{Setup}
Following \cite{li2024flona}, we conduct our experiments in the Gibson simulator \cite{xia2018gibson}. To ensure consistent evaluation across different methods and mitigate potential failures caused by environment mesh imperfections, both discrete and continuous action outputs are converted into global positions and orientations for execution. At inference, \policyname predicts the next $16$ waypoints, and the first $10$ are selected as the next action sequence.
Additionally, to reduce stochasticity, \policyname and its variants generate $30$ trajectories per inference step and average them to produce the next action sequence.
We employ a safety mechanism to mitigate collisions. Specifically, if a collision occurs, the agent returns to its previous state, reorients itself toward the goal, and then infers the following action based on the updated state. 
We record the collision events throughout the episode. An episode is considered successful if the agent reaches within a distance $d$ of the goal within at most $T_{\max}$ steps and incurs fewer than $\delta_{\max}$ collisions. In our experiments, we set $d=1$, $T_{\max}=500$, and $\delta_{\max}=15$.
\paragraph{Metrics} We evaluate performance using Success Rate (SR) and Success Weighted by Path Length (SPL). In ObjectNav, multiple instances of the same category may exist within a scene. An episode is deemed successful once the agent reaches any instance of the target category, while SPL is computed relative to the shortest geodesic path from the starting position to the specific instance encountered.

\subsection{Impact of Floor-Plan Priors}
\label{subsec: impact of floorplan}
To assess the floor plan’s impact across PointNav, ImageNav, and ObjectNav, we train a variant of the policy, denoted \policyname \textbf{w/o F}, from scratch, which removes the floor plan from the input while keeping all other components identical to \policyname. As shown in \cref{tab:w/o F}, removing the floor plan consistently reduces performance: SR drops by $16.2\%$, $6.3\%$, and $7.7\%$, and SPL decreases by $11.0\%$, $4.0\%$, and $5.8\%$ for PointNav, ImageNav, and ObjectNav, respectively. Note that because Gibson $4$+ lacks image and object goals (\cref{tab:datasets}), ImageNav and ObjectNav are evaluated only in HM3D, whereas PointNav is evaluated in both Gibson $4$+ and HM3D.

\begin{table}[t!]
    \centering
    \caption{\textbf{Comparison between \policyname and \policyname \textbf{w/o F}.}}
    \label{tab:w/o F}
    \resizebox{\linewidth}{!}{%
        \begin{tabular}{ccccccc}% 
            \toprule
            \multicolumn{1}{c}{\multirow{2}[1]{*}{Method}} & \multicolumn{2}{c}{PointNav} & \multicolumn{2}{c}{ImageNav} & \multicolumn{2}{c}{ObjectNav} \\
            \cmidrule{2-3}\cmidrule(lr){4-5}\cmidrule{6-7} & \multicolumn{1}{c}{SR} & \multicolumn{1}{c}{SPL} & \multicolumn{1}{c}{SR} & \multicolumn{1}{c}{SPL} & \multicolumn{1}{c}{SR} & \multicolumn{1}{c}{SPL} \\ 
            \midrule
            \policyname \textbf{w/o F}  & $25.8$      & $25.6$      & $22.6$      & $18.4$      & $20.9$      & $16.5$      \\
            \policyname                 & \bm{$42.0$} & \bm{$36.6$} & \bm{$28.9$} & \bm{$22.4$} & \bm{$28.6$} & \bm{$22.3$} \\ 
            \bottomrule
        \end{tabular}%
    }%
\end{table}

The consistent reduction in both SR and SPL indicates that \policyname effectively leverages floor-plan spatial priors to achieve more reliable and more efficient navigation.
The most significant improvement appears in PointNav, where \policyname outperforms \policyname\textbf{ w/o F} by $16.2\%$ in SR and $11.0\%$ in SPL. This substantial gain arises because PointNav benefits directly from the relatively explicit geometric priors provided by the floor plan, which guide the agent toward a precise goal location in previously unseen environments and significantly reduce navigation uncertainty.
In contrast, ImageNav and ObjectNav show minor improvements. For these tasks, the floor plan offers only coarse semantic cues—such as room layout and spatial organization—rather than explicit object- or image-level information. While the policy can make limited use of these cues, their indirect nature limits their contribution to the required semantic reasoning, yielding more modest gains than in PointNav.

\begin{figure*}[t!]
    \centering
    \includegraphics[width=\linewidth]{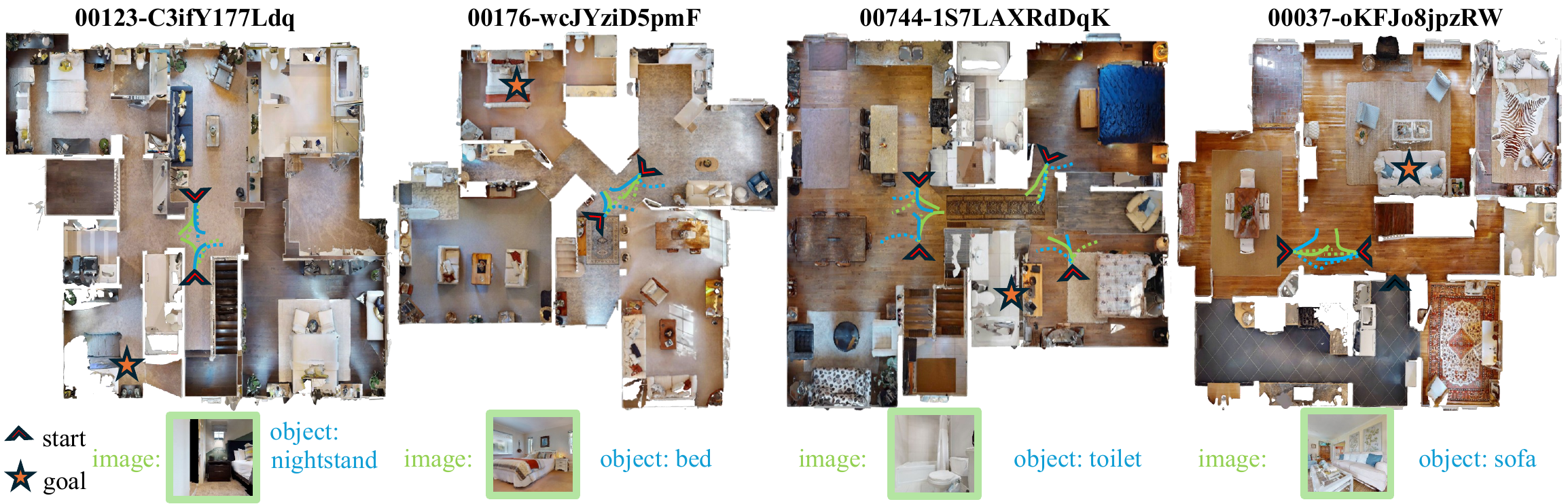}
    \caption{\textbf{Qualitative results demonstrating the guiding effect of the floor plan.}Top row: Across four HM3D scenes, we define a goal location and multiple starting points. The initial trajectories of \policyname (solid lines) and its variant without floor plan input (dashed lines) are shown under \textcolor{softlime}{image-goal} and \textcolor{cyan}{object-goal} conditions.}
    \label{fig:wo F}
\end{figure*}

As depicted in \cref{fig:wo F}, we conduct additional qualitative experiments to examine how floor-plan priors guide planning in ImageNav and ObjectNav. We visualize the mean initial predicted trajectories in four scenes—00123-C3ifY177Ldq, 00176-wcJYziD5pmF, 00744-1S7LAXRdDqK, and 00037-oKFJo8jpzRW. In each scene, an object is selected as the target, its rendered view is used as the image goal, and multiple starting points are randomly sampled. 
Across scenes, \policyname tends to produce trajectories that move more directly toward the goal, while those from \policyname \textbf{w/o F} exhibit greater variance and more frequent exploratory movements.
This contrast indicates that incorporating floor plan priors—such as coarse room layout cues—may help the agent better infer goal locations and improve navigation efficiency in ImageNav and ObjectNav.

\begin{table}[t!]
    \centering
    \caption{\textbf{Comparison between \policyname trained on individual modalities and jointly trained with random modality masking.}}
    \label{tab: masking vs single}
    \resizebox{\linewidth}{!}{%
        \begin{tabular}{ccccccc}%
            \toprule
            \multicolumn{1}{c}{\multirow{2}[1]{*}{Method}} & \multicolumn{2}{c}{PointNav} &  \multicolumn{2}{c}{ImageNav} & \multicolumn{2}{c}{ObjectNav} \\
            \cmidrule{2-3}\cmidrule(lr){4-5}\cmidrule{6-7} & \multicolumn{1}{c}{SR} & \multicolumn{1}{c}{SPL} & \multicolumn{1}{c}{SR} & \multicolumn{1}{c}{SPL} & \multicolumn{1}{c}{SR} & \multicolumn{1}{c}{SPL} \\ 
            \midrule
            Point-only   & \bm{$42.1$} & \bm{$38.9$} & -      & -      & -      & -      \\
            Image-only   & -      & -      & $25.4$ & $20.7$ & -      & -      \\
            Object-only  & -      & -      & -      & -      & $24.3$ & $19.9$ \\
            \policyname  & $42.0$ & $36.6$ & \bm{$28.9$} & \bm{$22.4$} & \bm{$28.6$} & \bm{$22.3$} \\
            \bottomrule
        \end{tabular}%
    }%
\end{table}

\subsection{Cross-Modality Complementarity}
\label{subsec: random masking}
To investigate the complementarity among different goal modalities, we conduct a comparative ablation study in which three single-modality policies—Point-only, Image-only, and Object-only—are trained separately, while \policyname is trained with all modalities using random masking. As shown in \cref{tab: masking vs single}, \policyname performs comparably with the Point-only model on PointNav but consistently outperforms all single-modality models on ImageNav and ObjectNav.

Interestingly, the Point-only policy achieves even marginally higher PointNav performance than \policyname, and it substantially outperforms the Image-only and Object-only models by over $15\%$ in SR and nearly $20\%$ in SPL. This pattern suggests that explicit geometric cues in the floor plan are easier to learn.

The improved ImageNav and ObjectNav results of \policyname further imply that the spatial knowledge learned from PointNav generalizes to semantic goal modalities, providing beneficial structural cues such as room connectivity and coarse spatial layout. 
In contrast, the slight drop in \policyname on PointNav indicates that the additional training signals from ImageNav and ObjectNav introduce more implicit, layout-level semantic biases in the encoder, which slightly interfere with the learning of explicit geometric structures required by PointNav.

\begin{figure}[t!]
    \centering
    \includegraphics[width=\linewidth]{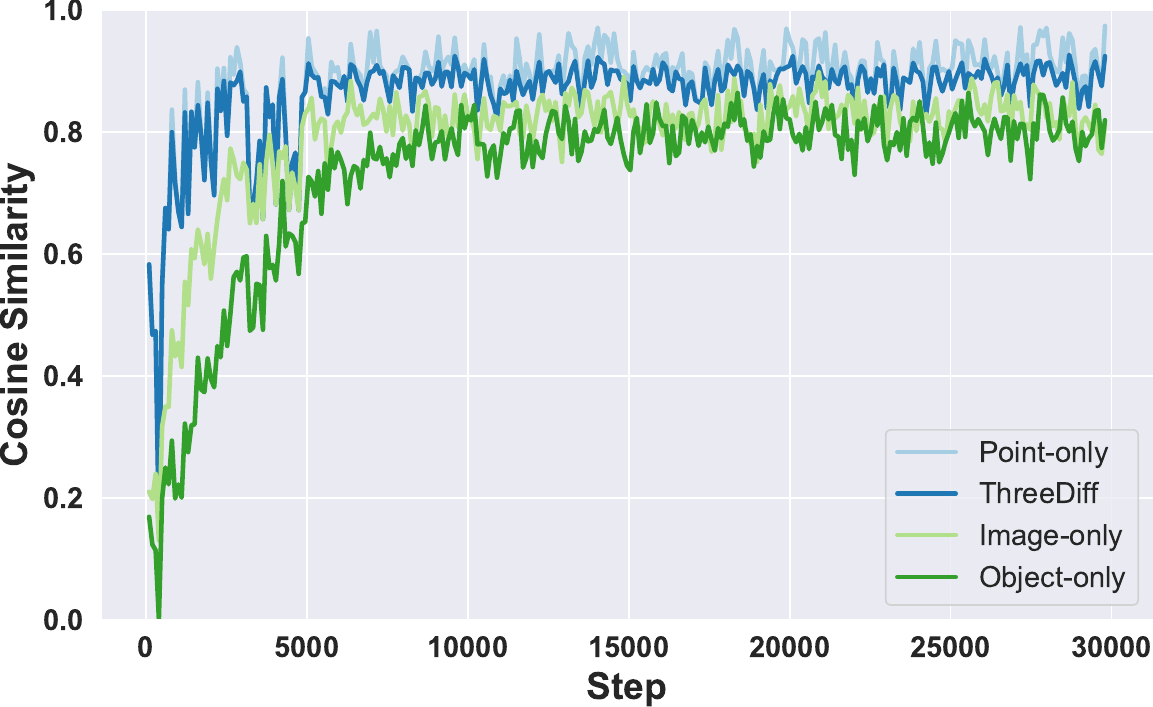}
    \caption{\textbf{Comparison of trajectory similarity changes during training.} Higher similarity indicates better model fitting ability. }
    \label{fig:cosine_similarity}
\end{figure}

Finally, \cref{fig:cosine_similarity} presents the evolution of cosine similarity between predicted and ground-truth trajectories for different training strategies. The curves show that \policyname and the Point-only model maintain similarly high alignment, both clearly outperforming Image-only and Object-only. This result again reflects that PointNav supervision drives the encoder to capture explicit geometric priors, whereas ImageNav and ObjectNav supervision encourage more implicit semantic layout understanding. Given that all models share the same floor-plan encoder, these results collectively highlight how different modalities emphasize complementary aspects of spatial representation.

\subsection{Comparison with Existing Models}
\label{subsec: comparison}
To provide a comprehensive and fair evaluation of \policyname, we compare it with several representative navigation methods. Specifically, we evaluate \textbf{DD-PPO} \cite{wijmans2020ddppo}, an RL model pretrained on Gibson, using its depth variant with a ResNet50 + LSTM512 backbone, without any additional fine-tuning. We also assess \textbf{ZSON} \cite{majumdar2022zson}, an ObjectNav model trained on HM3D, directly using its released weights.
In addition, we benchmark two imitation learning approaches—\textbf{Nomad} \cite{sridhar2023nomad} and \textbf{FloDiff} \cite{li2024flona}. Since both are originally RGB-only, we augment them with the same depth encoder used in \policyname and fine-tune their pre-trained checkpoints on our dataset to ensure a fair comparison.
Furthermore, we include three multi-modal navigation models from \cite{khanna2024goat}, enabling a direct comparison with existing multi-modality baselines.

\begin{table}[t!]
    \centering
    \caption{\textbf{Performance of baselines on the PointNav task in the Gibson $4$+ and HM3D scenes.}}
    \label{tab:baseline-point}
    % \resizebox{\linewidth}{!}{%
        \begin{tabular}{ccccc}%
            \toprule
            \multicolumn{1}{c}{\multirow{2}[1]{*}{Method}} & \multicolumn{2}{c}{Gibson $4$+} & \multicolumn{2}{c}{HM3D} \\
            \cmidrule{2-3}\cmidrule(lr){4-5} & \multicolumn{1}{c}{SR} & \multicolumn{1}{c}{SPL}  & \multicolumn{1}{c}{SR} & \multicolumn{1}{c}{SPL} \\
            \midrule
            DD-PPO \cite{wijmans2020ddppo}           & $18.3$      & $10.5$      & $13.7$      & $7.0$       \\
            FloDiff (pretrain)\cite{li2024flona}    & $40.0$      & $28.8$      & $23.7$      & $17.8$      \\
            FloDiff (finetune)\cite{li2024flona}    & $38.8$      & $29.6$      & $27.7$      & $21.1$      \\
            \policyname                             & \bm{$54.4$} & \bm{$50.0$} & \bm{$38.1$} & \bm{$32.5$} \\
            \bottomrule
        \end{tabular}%
    % }%
\end{table}

As shown in \cref{tab:baseline-point}, DD-PPO achieves an $18.3\%$ success rate in the Gibson $4$+ scenes (compared to $95.6\%$ SPL on the Habitat Challenge $2019$ \cite{savva2019habitat} validation set), despite being trained on all of these scenes. This drop is mainly due to task differences: unlike prior simpler settings that ignore collisions with scene objects, our task emphasizes obstacle avoidance and safer navigation—capabilities that DD-PPO does not possess. 
FloDiff (pre-trained) achieves success rates of $40\%$ in Gibson $4$+ and $23.7\%$ in HM3D, owing to its integration of floor plan information, which enhances its planning capabilities. After fine-tuning, FloDiff shows a slight drop on Gibson $4$+ but gains $4\%$ on HM3D, indicating that incorporating depth information can further improve performance. \policyname attains the best overall results, thanks to two key factors: (1) it has learned richer associations between multiple modalities and floor plans, leading to stronger spatial reasoning and navigation planning, and (2) it exhibits enhanced obstacle awareness, benefiting from the second-stage trajectory refinement (see \cref{subsection: second stage}).

As shown in \cref{tab:baseline-img&obj}, on ImageNav, \policyname outperforms Nomad (both pretrained and fine-tuned) in success rate and SPL, achieving $6.1\%$ higher SR and $5.4\%$ higher SPL than the fine-tuned version, despite both being trained on the same expert trajectories. This indicates that incorporating floor-plan priors facilitates more reliable and efficient image navigation. On ObjectNav, ZSON performs poorly, mainly because its training episodes are short (under $5\mathrm{m}$), while most episodes in our test set exceed $7\mathrm{m}$, limiting its ability to handle long-range goal inference.

\begin{table}[t!]
    \centering
    \caption{\textbf{Evaluation on the ImageNav and ObjectNav tasks.}}
    \label{tab:baseline-img&obj}
    % \resizebox{\linewidth}{!}{%
        \begin{tabular}{ccccc}%
            \toprule
            \multicolumn{1}{c}{\multirow{2}[1]{*}{Method}}  & \multicolumn{2}{c}{ImageNav} & \multicolumn{2}{c}{ObjectNav} \\
            \cmidrule{2-3}\cmidrule(lr){4-5} & \multicolumn{1}{c}{SR} & \multicolumn{1}{c}{SPL} & \multicolumn{1}{c}{SR} & \multicolumn{1}{c}{SPL} \\
            \midrule
            ZSON \cite{majumdar2022zson}                  & -             & -             & $7.2$       & $1.1$  \\
            NoMad (pretrain) \cite{sridhar2023nomad}      & $15.4$       & $10.9$       &   -          &  -    \\
            NoMad (finetune) \cite{sridhar2023nomad}      & $22.8$       & $17.0$       &   -          &  -      \\
            \policyname                                   & \bm{$28.9$}  & \bm{$22.4$}  & \bm{$28.6$} & \bm{$22.3$}\\
            \bottomrule
        \end{tabular}%
    % }%
\end{table}

\begin{table}[t!]
    \centering
    \caption{\textbf{Results of \policyname on the GOAT-Bench \cite{khanna2024goat}.} We compare \policyname with the three official baselines.}
    \label{tab:goat}
    % \resizebox{\linewidth}{!}{%
        \begin{tabular}{ccccc}%
            \toprule
            \multicolumn{1}{c}{\multirow{2}[1]{*}{Method}} & \multicolumn{2}{c}{ImageNav} & \multicolumn{2}{c}{ObjectNav} \\
            \cmidrule{2-3}\cmidrule(lr){4-5} & \multicolumn{1}{c}{SR} &\multicolumn{1}{c}{SPL} & \multicolumn{1}{c}{SR} & \multicolumn{1}{c}{SPL} \\
            \midrule
            RL Monolithic \cite{khanna2024goat}  & $11.7$          & $7.3$           & $25.7$          & $13.7$         \\
            RL Skill Chain\cite{khanna2024goat}  & \bm{$42.2$}     & $18.0$          & $25.8$          & $17.0$         \\
            Modular Goat\cite{khanna2024goat}    & $27.9$          & $19.5$          & $29.4$          & $17.0$         \\
            \policyname                          & $31.0$          & \bm{$25.6$}     & \bm{$34.5$}     & \bm{$30.8$}    \\
            \bottomrule
        \end{tabular}%
    % }%
\end{table}

We further compare \policyname with RL Monolithic, RL Skill Chain, and Modular GOAT \cite{khanna2024goat}. To enable direct comparison, we evaluate \policyname on the val-seen split of GOAT-Bench \cite{khanna2024goat}. As shown in \cref{tab:goat}, on ImageNav, \policyname attains a lower success rate than RL Skill Chain but achieves a $7.6\%$ higher SPL, indicating more efficient path planning. On ObjectNav, \policyname outperforms all three baselines in both success rate and SPL, demonstrating that leveraging floor-plan spatial information facilitates more reliable and efficient navigation.

It is worth noting that Modular GOAT, a map-based approach, achieves higher SPL than both RL Monolithic and RL Skill Chain, but still falls short of \policyname by $6.1\%$ on ImageNav and $13.8\%$ on ObjectNav. While explicit mapping aids in preserving environmental memory and supports long-horizon planning, it requires additional movements during initial exploration, which reduces overall efficiency. In contrast, \policyname exploits spatial priors in the floor plan, including both geometric structures and semantic regularities, providing guidance that reduces unnecessary exploration and enables more reliable and efficient navigation across tasks.

\subsection{Ablation of Refiner}
\label{subsection: second stage}
\begin{figure}[t!]
    \centering
    \includegraphics[width=\linewidth]{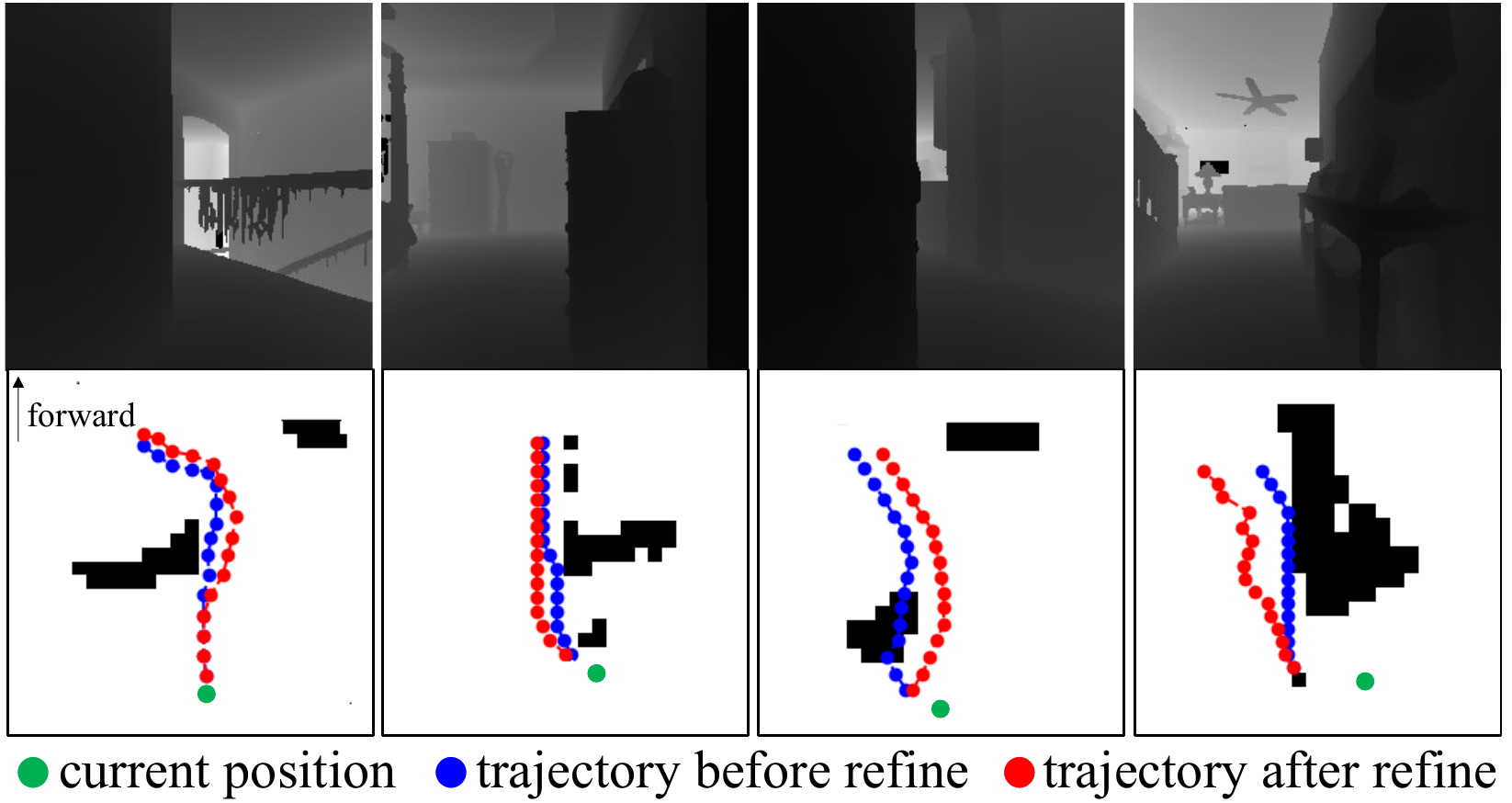}
    \caption{\textbf{Impact of the second-stage refinement.} Top row: depth observation at an inference step. Bottom row: corresponding local occupancy map with trajectories before and after refinement.}
    \label{fig:refiner}
\end{figure}

To assess the impact of the refiner, we ablate the second-stage module, producing the w/o refiner variant. As reported in \cref{tab:ablation on refiner}, w/o refiner incurs performance degradation across all tasks, with the most significant decline observed on PointNav. This underscores the critical role of the second-stage refinement in enhancing overall trajectory quality. \cref{fig:refiner} provides a visual comparison of trajectories before and after refinement. The refined trajectory clearly avoids obstacles, demonstrating that by converting depth information into local SDF, the model effectively captures navigability cues and generates safer waypoints.

\begin{table}[t!]
    \centering
    \caption{\textbf{Performance of \textbf{w/o refiner} and \policyname.}}
    \label{tab:ablation on refiner}
    \resizebox{\linewidth}{!}{%
        \begin{tabular}{ccccccc} 
            \toprule
            \multicolumn{1}{c}{\multirow{2}[1]{*}{Method}} & \multicolumn{2}{c}{PointNav} & \multicolumn{2}{c}{ImageNav} & \multicolumn{2}{c}{ObjectNav} \\
            \cmidrule{2-3}\cmidrule(lr){4-5}\cmidrule{6-7} & \multicolumn{1}{c}{SR} & \multicolumn{1}{c}{SPL} & \multicolumn{1}{c}{SR} & \multicolumn{1}{c}{SPL} & \multicolumn{1}{c}{SR} & \multicolumn{1}{c}{SPL} \\ 
            \midrule 
            \textbf{w/o refiner}  & $34.3$      & $29.2$      & $27.7$      & $21.4$      & $25.3$      & $19.4$      \\ 
            \policyname           & \bm{$42.0$} & \bm{$36.4$} & \bm{$28.9$} & \bm{$22.4$} & \bm{$28.6$} & \bm{$22.3$} \\ 
            \bottomrule
        \end{tabular}
    }%
\end{table}

\section{Conclusion}
We present \taskname, a novel task that challenges agents to perform multimodal goal-directed navigation using floor plans and egocentric RGBD observations. To support research on this task, we collect \datasetname, a large-scale dataset suitable for both training and evaluation. Furthermore, we introduce \policyname, a two-stage end-to-end model capable of effectively handling multiple navigation modalities. Comprehensive quantitative and qualitative assessments demonstrate (1) floor plans significantly improve navigation efficiency, and (2) \policyname achieves robust performance across all three navigation tasks. We believe our work will serve as a solid foundation for future advances in visual navigation.

\paragraph{Limitations}
\policyname has two limitations. First, it has been primarily evaluated on floor plans from residential indoor environments, and its performance on other types of layouts remains to be explored.
Second, it has not yet been tested in real-world scenarios, leaving its practical deployment to be investigated in future work.

\paragraph{Acknowledgments}
This project is supported by Beijing Natural Science Foundation (L242094) and the National Natural Science Foundation of China (NSFC, 62172043).

\small
\bibliographystyle{ieeenat_fullname}
\bibliography{reference}
% \clearpage
% \setcounter{page}{1}
% \maketitlesupplementary
% \renewcommand{\thesection}{\Alph{section}}
% \setcounter{section}{0}
\clearpage
\appendix
\renewcommand\thefigure{A\arabic{figure}}
\setcounter{figure}{0}
\renewcommand\thetable{A\arabic{table}}
\setcounter{table}{0}
\renewcommand\theequation{A\arabic{equation}}
\setcounter{equation}{0}
\pagenumbering{arabic}% resets `page` counter to 1
\renewcommand*{\thepage}{A\arabic{page}}
\setcounter{footnote}{0}
\maketitlesupplementary

\section{Floor Plan Construction}
\label{sec:Floor Plan Construction}
\vspace*{-0.5\baselineskip}
\begin{algorithm}[h!]
    \caption{Floor Plan Extraction}
    \label{alg:floorplan}
    \KwIn{3D scene in OBJ format}
    \KwOut{floor plan $F \in \mathbb{R}^{H_f \times W_f}$} %, where wall pixels = 0 and empty space = 255 \\
    \begin{algorithmic}[1]
        % \STATE \textbf{// Load Mesh}
        % \STATE $M \leftarrow$ triangulated faces in OBJ
        
        % \STATE \textbf{// Determine Floor Heights}
        % \STATE $M_{\text{v}} \leftarrow$ faces in $M$ whose normals are nearly vertical.
        % \STATE $\textit{heights} \leftarrow$ z-coordinates of faces in $M_{\text{v}}$
        % \STATE $\textit{clusters} \leftarrow$ cluster the heights 
        % \STATE $h_\text{floor} \leftarrow$ the cluster centroid with the largest support
        % \STATE \textbf{// Horizontal slicing to extract floor plans}
        % \STATE $\textit{cutting\_h} \leftarrow \{\, h_\text{floor} + 1.25  \}$ 
        % \STATE $\textit{cut\_lines} \leftarrow$ intersection lines between each horizontal plane at heights in $\textit{cutting\_h}$ and all faces in $M$
        % \STATE $\textit{contours} \leftarrow$ connect line segments in $\textit{cut\_lines}$ that share endpoints using BFS to form wall polylines
        % \STATE $F^* \leftarrow$ render \textit{contours} onto an image canvas
        % \STATE \textbf{// Post-processing}
        % \STATE $F^* \leftarrow$ apply median filter to remove small noise
        % \STATE $F^* \leftarrow$ apply morphological dilation and erosion
        % \STATE $F \leftarrow$ apply median filter to smooth wall boundaries
        
        % \vspace{0.3em}
        % \STATE \textbf{// Notation} 
        % \STATE $\mathbf{n}(f)$: outward unit normal of face $f$
        % \STATE $z(f)$: z-coordinate of the centroid of face $f$

        % \vspace{0.3em}
        \STATE \texttt{// Floor Height Estimation}
        \STATE $M_{\mathrm{v}} \gets \{\, f \in M \mid \mathbf{n}(f) \text{ is nearly vertical} \,\}$
        \STATE $\textit{H} \gets \{\, z(f) \mid f \in M_{\mathrm{v}} \,\}$
        \STATE $\textit{C} \gets \textsc{Cluster}(\textit{H})$
        \STATE $h_{\mathrm{floor}} \gets \textsc{Centroid}(\arg\max_{c \in \textit{C}} |c|)$
        
        % \vspace{0.3em}
        \STATE \texttt{// Horizontal Slicing for Floor Plan Extraction }
        \STATE $\textit{h\_cut} \gets \{\, h_{\mathrm{floor}} + 1.25 \,\}$
        \STATE $\textit{s\_cut} \gets \textsc{Intersect}(M, \textit{h\_cut})$
        \STATE $\textit{cts} \gets \textsc{TraceContours}(\textit{s\_cut})$
        
        % \vspace{0.3em}
        \STATE $\tilde{F_0} \gets \textsc{AssembleFloorPlan}(\textit{cts})$

        % \vspace{0.3em}
        \STATE \texttt{// Post-Processing }
        \STATE $\tilde{F_1} \gets \textsc{Denoising}(\tilde{F_0})$ \quad // remove small artifacts 
        \STATE $\tilde{F_2} \gets \textsc{Morphology}(\tilde{F_1};\, \text{dilate} \rightarrow \text{erode})$
        \STATE $F \gets \textsc{Denoising}(\tilde{F_2})$ \quad // smooth final boundaries
        \RETURN $F$
    \end{algorithmic}
\end{algorithm}
% Given the absence of floor plan annotations in many indoor datasets, we reconstruct 2D floor plans directly from 3D meshes using the pipeline in \cref{alg:floorplan}. We first detect vertically oriented faces and cluster their heights to estimate the floor level, then extract wall contours by slicing the scene with a horizontal plane placed at a fixed offset above the floor.

Given the lack of floor plan annotations in many indoor datasets, we reconstruct 2D floor plans from 3D scene meshes using the pipeline described in \cref{alg:floorplan}. 
We identify near-vertical faces by thresholding the angular deviation between face normals $\mathbf{n}(f)$ and the gravity direction, and cluster their centroid heights $z(f)$ to estimate the floor level. We then extract wall contours by intersecting the mesh with a horizontal plane at a fixed offset above the estimated floor.

The slicing offset is a critical design choice. Based on empirical analysis, we set the offset to $1.25\,\mathrm{m}$. 
Smaller offsets ($0.25$–$0.5\,\mathrm{m}$) intersect furniture and introduce significant high-frequency noise, while larger offsets (e.g., $2.0\,\mathrm{m}$) fail to capture continuous wall structures, leading to fragmented contours. 
The $1.25\,\mathrm{m}$ offset balances noise suppression and wall completeness, resulting in stable, coherent intersections.

Finally, we apply a morphological closing to consolidate fragmented wall segments, using $15$ dilation and $5$ erosion iterations with $13\times1$ and $1\times13$ kernels. We further suppress noise by removing small black connected components ($<100$ pixels) and holes enclosed by components smaller than $400$ pixels, yielding clean and consistent floor plan boundaries. The resulting floor plans are exported as PNG images, where wall structures are encoded in black. Moreover, \datasetname provides precise correspondence between floor plan pixel coordinates and the associated real-world coordinates for each scene.

\section{Navigable Map}
\begin{figure}[t]
    \centering
    \includegraphics[width=\linewidth]{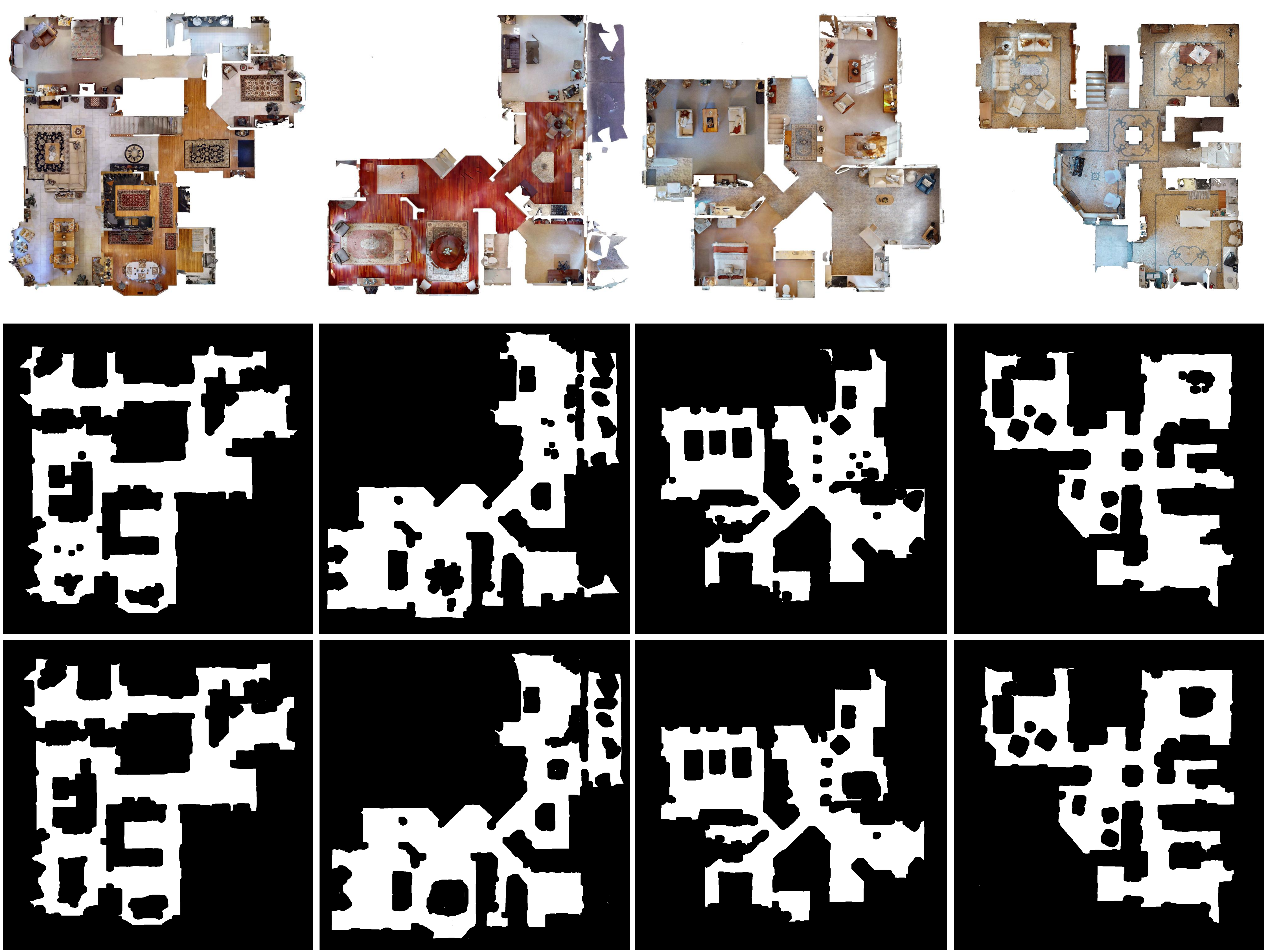}
    \caption{\textbf{High-Fidelity Navigable Maps in \datasetname.} Top row: Top-down views for the scene. Second row: navigable maps before manual annotation. Bottom row: navigable maps after manual annotation.}
    \label{fig:trav_map_compare}
\end{figure}
To enable robust imitation learning, we require collision-free trajectories grounded in accurate and consistent navigable maps. 
Although HM3D and Gibson provide navigability annotations, their quality and formats are inconsistent. 
We therefore reconstruct a unified navigable map for each scene and perform careful manual refinement to ensure dataset-wide reliability and consistency.

We construct the navigable map by projecting 3D mesh faces within a selected height range onto the ground plane and marking occupied regions as non-navigable. 
Heights that are too low capture floor reconstruction noise, while excessive heights (e.g., door frames and overhanging structures) are mistakenly treated as obstacles, fragmenting free space. 
Based on empirical validation, we set the height range to $0.3$–$2.0\,\mathrm{m}$.

Finally, we manually verify and correct all generated maps to handle missing or spurious mesh artifacts arising from reconstruction errors. 
The resulting navigable maps are produced at a resolution of $1\,\mathrm{cm}$/pixel, as shown in \cref{fig:trav_map_compare}.

\begin{table*}[t]
\centering
\caption{Object categories in \datasetname.}
\label{tab:object}
\renewcommand{\arraystretch}{1.2}
\begin{tabular}{p{0.95\textwidth}}
\hline
\textit{
sink, blanket, bathroom cabinet, cabinet, chair, table, rack, tv, bathroom shelf, window, couch, stove, bed, toilet, support beam, shelving, wardrobe, sideboard, easy chair, kitchen lower cabinet, desk, bench, sofa set, stool, tray, board game, storage, screen, radiator, washing machine, ladder, dining\_chair, sofa, washing\_machine, dining\_table, shower, bathroom\_cabinet, nightstand, cupboard, bookcase, case, sign, box, sheet, treadmill, bag, rug, washer-dryer, dishwasher, pillow, blinds, mirror, bathtub, cooker, bathtub platform, refrigerator, plant, sink cabinet, mantle, dressing\_table, towel, shower cabin, pot, kitchen cabinet lower, crate, pillar, hat, clothes, oven, shower bar, board, heater, antique clock, handbag, desk lamp, radio, curtain valence, kitchen appliance, counter, display cabinet, figure, flag, book rack, hunting trophy, cloth, wine\_cabinet, plants, carpet, oven and stove, storage cabinet, vacuum cleaner, bath towel, cabinet clutter, record player, air conditioner, stovetop, paper, coffee\_table, side\_table, cardboard box, clutter, pouffe, laundry, wood, basket, brochure, note, office chair, canvas, banner, drawer sink table, duct, drawer, sofa seat, closet shelving, scarf, kitchen top, shower rail, bathrobe, bathroom towel, tv\_cabinet, book, picture, footstool, monitor, window shade, countertop, bottle, laundry basket, shelf cubby, vase, microwave, flower vase, shower glass, clothes dryer, whiteboard, washbasin counter, kitchen table, bulletin board, island, stair, bedframe, dresser, plush toy, window shutter, sofa chair, cardboard, stack of papers, jacket, toy, guitar, flowerpot, closet, piano, bar\_chair, worktop, shower tap, chimney, washbasin, sink table, candle, lounge chair, storage shelving, plate, freezer, boiler, folding table, trampoline, container, water tank, gas furnace, sleeping bag, electrical controller, patio chair, grill, rocking chair, stereo set, closet shelf, seat, lamp stand, gym equipment, dish rack, crib, cradle, bathroom accessory, tool, bucket, shower soap shelf, paper towel, spice rack, antique telephone, tub, light switch, jewelry box, potted\_bonsai, coffee machine, bowl of fruit, glass, clock, dinner table, bar, hose, ottoman, display table, grass, shower-bath cabinet, decorative quilt, trashcan, magazine, stand, binder, amplifier, hutch, dressing table, kitchen extractor, entrance\_cabinet, balustrade, telescope, sunbed, clothing stand, paper storage, dish cabinet, candle holder, kitchen shelf, elevator, office table, window glass, iron board, handrail, printer, laptop, tent, ornament, handle, soft chair, table stand, globe, decorative\_cabinet, shoe\_cabinet, aquarium, kitchen sink, backrest, stage, massage bed, exercise ladder, bed curtain, locker, pool, bed comforter, headboard, coat, cabinet table, parapet, exercise bike, photo, stack of stuff, desk cabinet, book cabinet, photo stand, shower stall, shade, water dispenser, archway, storage space, fire extinguisher, exhibition panel, solarium, pile of magazines, cosmetics, kitchen countertop items, statue, calendar, lampshade, hook, painting frame, baby changing station, purse, tile, artwork, arcade game, backsplash, fish tank, pantry, newspaper, cutting board, speaker, bunk bed, jar, decorative plant, floor-standing\_lamp, telephone, dishrag, washing\_cabinet, pool table, hanging clothes, umbrella, cart, barbecue, hanger, foosball game table, high shelf, range hood, robe, hammock, bar cabinet, clothes bag, violin case, round chair, folding stand, wreath, spa bench, tank
}\\
\hline
\multicolumn{1}{c}{(a) Object Categories in the Training Set} \\[6mm]
\hline
\textit{
lounge chair, bench, bed, aquarium, double armchair, cabinet, medical lamp, washbasin, chair, sauna oven, brochure, foot spa, pot, box, towel, relief, window, massage bed, seat, backrest, table, laptop, sign, basket, crate, desk, toilet, nightstand, bathroom\_cabinet, picture, refrigerator, stair, dishwasher, rug, sink, couch, tv, wardrobe, display cabinet, bar, heater, oven and stove, blanket, kitchen top, pillar, dining\_table, dining\_chair, sideboard, cooker, sink cabinet, side\_table, sofa, ladder, shower-bath cabinet, cardboard box, vacuum cleaner, board, statue, toy, carpet, stool, dressing\_table, shower, stove, worktop, office chair, washing machine, clothes dryer, painting frame, handle, plants, floor-standing\_lamp, oven, jacket, clothes, decorative quilt, countertop, bathtub, pouffe, decorative plate, closet, rocking chair, rack, shower tap, bathroom cabinet, stand, bathroom shelf, bath towel, kitchen table, bathroom towel, container, clutter, stovetop, case, hat, ornament, pillow, staircase handrail, parapet, mirror, window glass, kitchen lower cabinet, glass, kitchen appliance, tv\_cabinet, coffee\_table, elevator, tub, high shelf, stone support structure, electrical controller, vase, kitchen shelf, storage shelving, calendar, island, flower vase, flowerpot, bar\_chair, gym equipment, balustrade, plush toy, decorative plant, radiator, shower cabin, display table, dinner table, telephone, fuse box, book, plant, window shade, microwave, drawer, photo mount, stack of papers, schedule, shade, shelving, cupboard, washing\_machine, shoe\_cabinet, mixer, wine cabinet, handbag
}\\
\hline
\multicolumn{1}{c}{(b) Object Categories in the Validation Set} \\[10mm]
\end{tabular}
\end{table*}
\noindent\mbox{}
\begin{figure*}[t!]
    \centering
    \includegraphics[width=\linewidth]{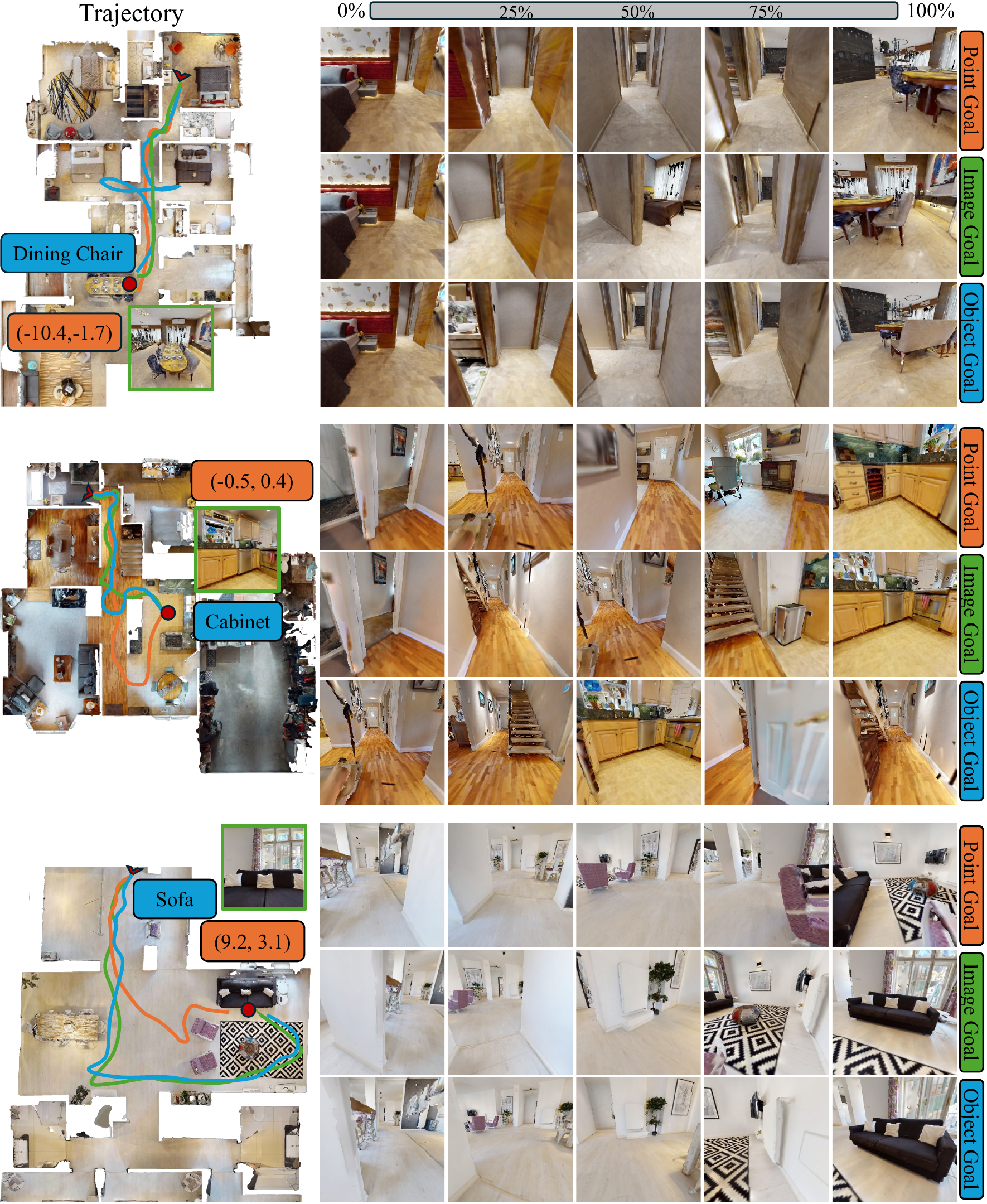}
    \caption{\textbf{Qualitative Results of \policyname on \datasetname.}}
    \label{fig:traj}
\end{figure*}

\section{Object Goal Annotation}
\begin{figure}[t]
    \centering
    \includegraphics[width=\linewidth]{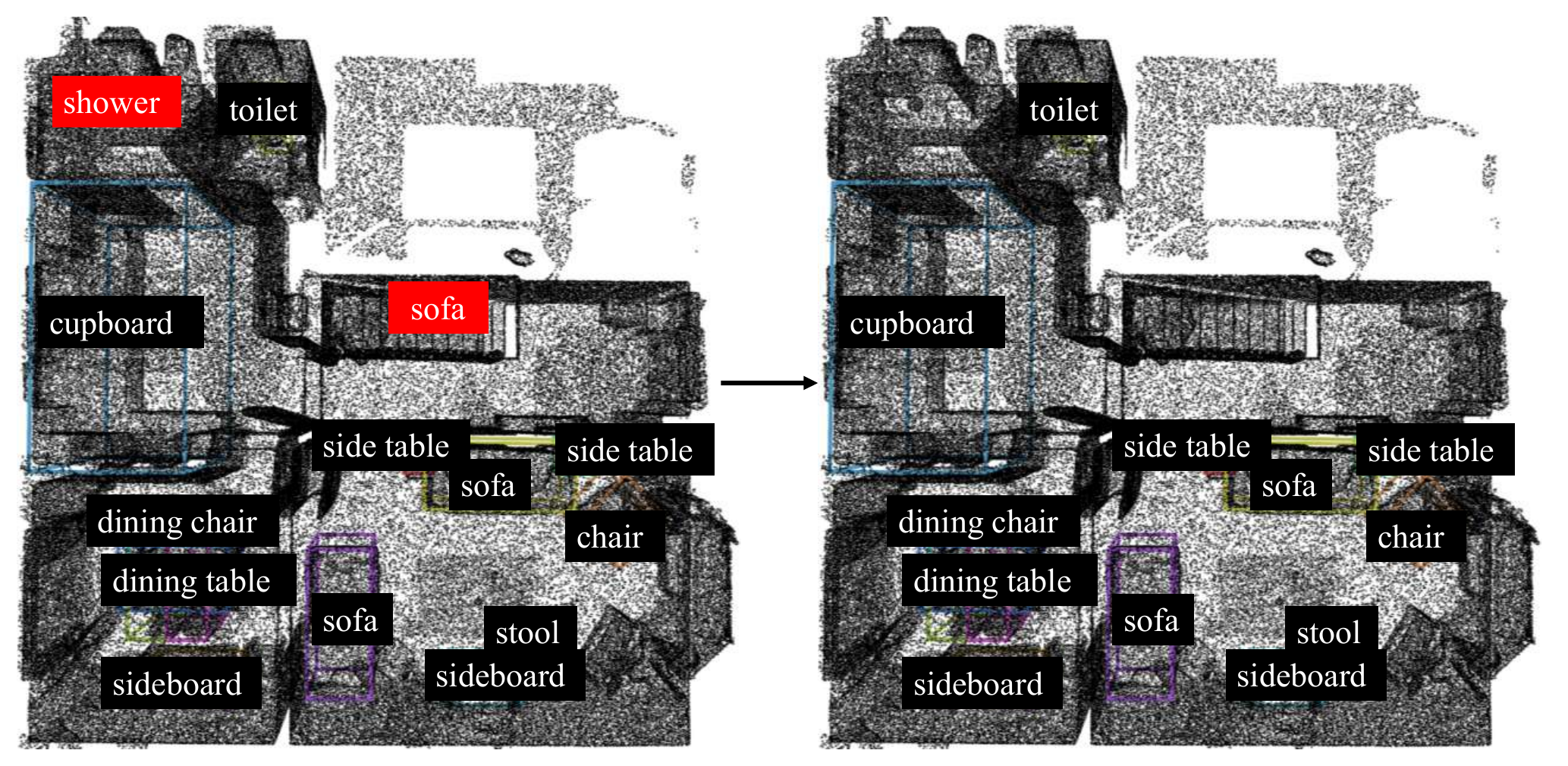}
    \caption{\textbf{The Manual Verification Process.} Left: detection results before manual verification, the red boxes indicate incorrect detections. Right: detection results after manual verification.}
    \label{fig:annotation}
\end{figure}

To enrich both the number of scenes with object-goal annotations and the diversity of object categories, we employ SpatialLM \cite{SpatialLM}, a 3D large language model designed to process 3D point cloud data and generate structured 3D scene understanding outputs. 
As SpatialLM is designed to operate on single-layer point clouds, we first decompose each scene into hierarchical layers following the procedure in \cref{sec:Floor Plan Construction}.
We then input each layer's point cloud into SpatialLM to obtain layer-wise object annotations. To prevent the outputs from containing incorrect detections, as shown in \cref{fig:annotation}, we manually verify and correct the predictions to ensure that all object annotations are accurate.

To ensure the geometric validity of object placements, we perform an additional verification on all objects. Specifically, we remove two categories of invalid placements: (1) objects whose centers lie within navigable regions, since all objects should fall in non-traversable areas; (2) objects in overly confined or cluttered regions, where their placement prevents feasible observation viewpoints. The complete set of object categories included in our dataset is summarized in \cref{tab:object}. 

\begin{table}[t!]
    \centering
    \caption{\textbf{Additional Experiments.} Comparison between finetuned baselines and \policyname.}
    % \vspace{-5pt}
    \label{tab:additional}
    \resizebox{\linewidth}{!}{%
        \begin{tabular}{ccccc}
            \toprule
            \multicolumn{1}{c}{\multirow{2}[1]{*}{Method}}  & \multicolumn{2}{c}{ImageNav} & \multicolumn{2}{c}{ObjectNav} \\
            \cmidrule{2-3}\cmidrule(lr){4-5} & \multicolumn{1}{c}{SR} & \multicolumn{1}{c}{SPL} & \multicolumn{1}{c}{SR} & \multicolumn{1}{c}{SPL} \\
            \midrule
            ZSON (finetune) \cite{majumdar2022zson}                  & -             & -             & $10.0$       & $9.1$  \\
            RL Monolithic (finetune) \cite{sridhar2023nomad}      & $8.4$       & $3.7$       &   $22.5$          &  $15.2$      \\
            \policyname                                   & \bm{$28.9$}  & \bm{$22.4$}  & \bm{$28.6$} & \bm{$22.3$}\\
            \bottomrule
        \end{tabular}
    }%
\end{table}

\section{Training Data Construction}
This section describes how we construct training samples for \policyname from expert trajectory data. For each waypoint sequence, we sample a starting point every four waypoints, and each selected start point is then paired with the final waypoint of the trajectory, forming multiple start–end pairs. For each pair, we extract the past $8$ waypoints as the historical context and the next $16$ waypoints as the future supervision signal. Formally, each training sample consists of $(S_h, P_{gt}, F, g)$, where the historical states $S_h = (s_{t-7},s_{t-6},...,s_{t},)$, and the ground-truth future positions $P_{gt} = (p_t, p_{t+1}, p_{t+2},..., p_{t+15})$. The goal specification $g$ is defined as $g = (g_{point}, g_{image}, g_{object})$ for IO episodes, and $g = g_{point}$ otherwise.

\section{Additional Experiments}
We further finetune the ZSON \cite{majumdar2022zson} and RL Monolithic \cite{khanna2024goat} models using the episodes in \datasetname. We train ZSON and RL Monolithic for an additional $80$M and $100$M steps, respectively, and then evaluate them on our validation set. As shown in \cref{tab:additional}, ZSON exhibits a modest performance gain after finetuning, consistent with expectations, yet it remains notably inferior to \policyname. In contrast, RL Monolithic experiences performance degradation, as illustrated in \cref{tab:goat}. This decline occurs because, unlike Goat-Bench, our evaluation protocol treats any episode with more than $15$ collisions as a failure, thereby imposing a stricter success criterion.

\section{More Qualitative Results}
We provide additional navigation examples of \policyname, as shown in the \cref{fig:traj}.

\section{Inference speed.}
On an NVIDIA 4090 GPU, \policyname runs at 0.18s/step, with each stage taking about 0.09s.

\end{document}